\documentclass[preprint,12pt,3p]{elsarticle}
\usepackage{framed,multirow}
\usepackage{float}
\usepackage{amssymb}
\usepackage{latexsym}
\usepackage{url}
\usepackage{color}
\usepackage{mathrsfs} 
\definecolor{inchworm}{rgb}{0.7, 0.93, 0.36} 
\definecolor{orng}{rgb}{1.0, 0.31, 0.0} 
\usepackage{algorithm}
\usepackage{algorithmic}
\usepackage{amsmath,amssymb,amsfonts}
\usepackage{graphicx}
\usepackage{textcomp}
\usepackage{cuted}
\usepackage{dblfloatfix} 
\usepackage{hyperref}

\hypersetup{colorlinks = true,
    linkcolor=blue,
    filecolor=cyan,
    citecolor = blue,
    urlcolor=blue}
\definecolor{ib}{rgb}{0.0, 0.18, 0.65}
\definecolor{fg}{rgb}{0.13, 0.55, 0.13}
\usepackage[english]{babel}
\addto\extrasenglish{

} 
\addto\captionsenglish{} 
\usepackage[labelfont=bf]{caption} 
\usepackage{relsize} 
\newcommand\eatpunct[1]{} 
\usepackage{booktabs} 
\usepackage[bb=boondox]{mathalfa} 
\usepackage{float} 
\usepackage{makecell} 
\usepackage{flushend} 
\usepackage[nameinlink,capitalize]{cleveref} 
\usepackage{setspace}

\usepackage{MnSymbol} 
\usepackage{mathrsfs} 
\usepackage{tabularx,booktabs} 
\newcolumntype{C}{>{\centering\arraybackslash}X} 
\usepackage{array}
\newcolumntype{P}[1]{>{\centering\arraybackslash}p{#1}} 
\newcolumntype{L}[1]{>{\arraybackslash}p{#1}} 
\usepackage{enumerate} 
\usepackage[labelformat=simple]{subcaption}
\journal{}

\begin{document}

\begin{titlepage}
\doublespacing
{\centering{\Large {MNet-SAt: A Multiscale Network with Spatial-enhanced Attention for Segmentation of Polyps in Colonoscopy}}\\
Chandravardhan Singh Raghaw, Aryan Yadav, Jasmer Singh Sanjotra, Shalini Dangi, Nagendra Kumar\\}

\vspace{2em}

{\large Highlights}

\begin{itemize}
    \item A novel edge-guided encoder–decoder framework to capture fine-grained edge features.
    \item Multiscale feature aggregator extracts diverse contextual features of polyps.
    \item Spatial enhanced attention captures spatial-aware global contextual features.
    \item Fusion of channel-wise multiscale features with inter-channel information.
    \item Quantitative and qualitative analyses demonstrate the superiority of MNet-SAt.
\end{itemize}

\vspace{2em}

\noindent This is the preprint version of the accepted paper.\\
\noindent This paper is accepted in \textbf{Biomedical Signal Processing and Control, 2025.}
\\
DOI: \url{https://doi.org/10.1016/j.bspc.2024.107363}
\end{titlepage}


\begin{frontmatter}
\title{MNet-SAt: A Multiscale Network with Spatial-enhanced Attention for Segmentation of Polyps in Colonoscopy}

\author[1]{Chandravardhan Singh Raghaw}
\ead{phd2201101016@iiti.ac.in}

\author[1]{Aryan Yadav}
\ead{cse200001010@iiti.ac.in}

\author[2]{Jasmer Singh Sanjotra}
\ead{ee220002041@iiti.ac.in}

\author[1]{Shalini Dangi}
\ead{phd2301201003@iiti.ac.in}

\author[1]{Nagendra Kumar\corref{cor1}}
\ead{nagendra@iiti.ac.in}
\cortext[cor1]{Corresponding author}

\address[1]{Department of Computer Science and Engineering, Indian Institute of Technology (IIT) Indore, Indore 453552, India}

\address[2]{Department of Electrical Engineering, Indian Institute of Technology (IIT) Indore, Indore 453552, India}

\begin{abstract}
\textit{Objective:} To develop a novel deep learning framework for the automated segmentation of colonic polyps in colonoscopy images, overcoming the limitations of current approaches in preserving precise polyp boundaries, incorporating multi-scale features, and modeling spatial dependencies that accurately reflect the intricate and diverse morphology of polyps.

\noindent\textit{Methods:} To address these limitations, we propose a novel Multiscale Network with Spatial-enhanced Attention (MNet-SAt) for polyp segmentation in colonoscopy images. This framework incorporates four key modules: Edge-Guided Feature Enrichment (EGFE) preserves edge information for improved boundary quality; Multi-Scale Feature Aggregator (MSFA) extracts and aggregates multi-scale features across channel spatial dimensions, focusing on salient regions; Spatial-Enhanced Attention (SEAt) captures spatial-aware global dependencies within the multi-scale aggregated features, emphasizing the region of interest; and Channel-Enhanced Atrous Spatial Pyramid Pooling (CE-ASPP) resamples and recalibrates attentive features across scales.

\noindent\textit{Results:} We evaluated MNet-SAt on the Kvasir-SEG and CVC-ClinicDB datasets, achieving Dice Similarity Coefficients of 96.61\% and 98.60\%, respectively.

\noindent\textit{Conclusion:} Both quantitative (DSC) and qualitative assessments highlight MNet-SAt's superior performance and generalization capabilities compared to existing methods.

\noindent\textit{Significance:} MNet-SAt's high accuracy in polyp segmentation holds promise for improving clinical workflows in early polyp detection and more effective treatment, contributing to reduced colorectal cancer mortality rates.
\end{abstract}

\begin{keyword}
Polyp segmentation \sep
Multi-scale features \sep
Spatial attention \sep
Encoder-Decoder \sep
Boundary preservation.
\end{keyword}

\end{frontmatter}

\section{Introduction}
\label{sec:introduction}
\begin{figure*}[!ht]
\captionsetup[subfigure]{justification=centering}
     \centering
     \begin{subfigure}[t]{0.3340\linewidth}\centering
         \includegraphics[width=\linewidth]{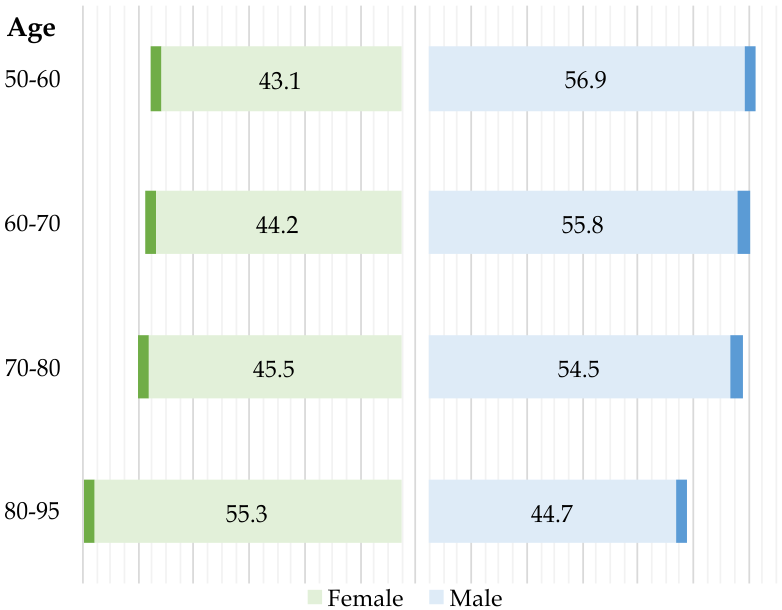}
         \caption{Age-Gender Distribution of Polyp Patients}\label{fig:intro-malefemale}
     \end{subfigure}\hfill
     \begin{subfigure}[t]{0.3898\linewidth}\centering
         \includegraphics[width=\linewidth]{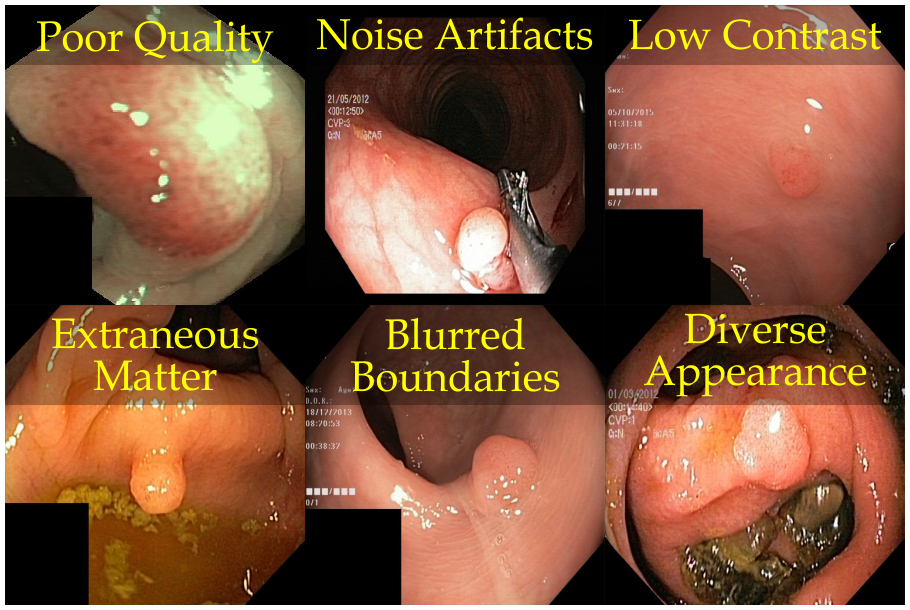}
         \caption{Challenges in Polyp Image Quality}\label{fig:intro-polyp}
     \end{subfigure}\hfill
     \begin{subfigure}[t]{0.2762\linewidth}\centering
         \includegraphics[width=\linewidth]{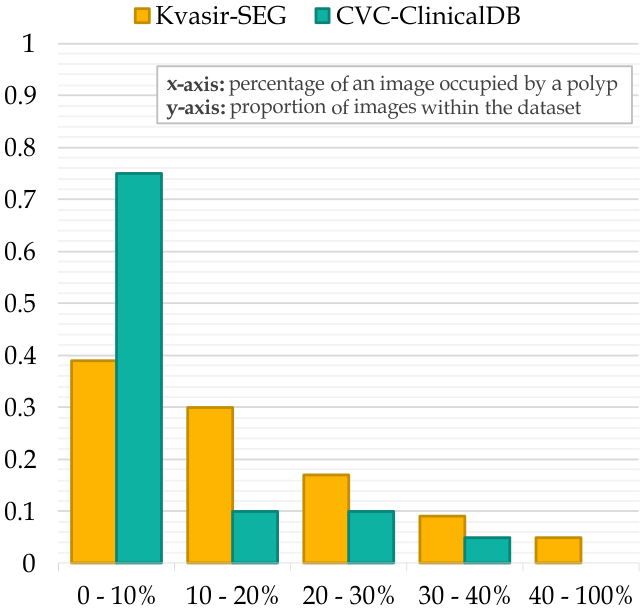}
         \caption{Polyp Size Distribution}\label{fig:intro-pixel}
     \end{subfigure}\hfill     
    \caption{Polyp patient demographics, image quality issues, and size distribution}
    \label{fig:intro}
\end{figure*}

Global health faces a significant burden from colorectal cancer (CRC), the second most commonly diagnosed cancer and third leading cause of cancer-related mortality~\cite {Siegel2024cancer}. Research indicates that approximately 95\% of CRC cases originate from colorectal adenomatous polyps~\cite{Rawla2019risk}. While the overall survival rate for CRC is 63\%, early detection and treatment at a localized stage can significantly improve this rate to 91\%~\cite{Rawla2019risk, Jiang2022pattern}. \cref{fig:intro-malefemale} summarize the statistics for polyps by age and gender. However, survival drops to 14\% when the cancer metastasizes to distant organs. Consequently, early polyp detection and treatment are crucial for CRC prevention and reducing mortality rates. However, the manual detection process is labor-intensive, and variability in clinician expertise can lead to overlooking polyps during colonoscopy.

Automated gastrointestinal polyp segmentation presents significant challenges due to various factors illustrated in~\cref{fig:intro-polyp}. Uneven illumination from body fluid reflection can degrade image quality, while noise artifacts like surgical instruments and intestinal contents complicate segmentation. Low contrast between tissues hinders accurate polyp localization and identification. Additionally, residual stool and digestive fluids can obscure internal tissues, making differentiation difficult. Polyp boundaries are blurred or unclear due to similar appearance patterns. Furthermore, inter-patient variability and diverse polyp types exhibit various colors and textures. Moreover, an imbalance in pixel distribution, as shown in~\cref{fig:intro-pixel}, with the background area often exceeding the polyp area, introduces noise that affects segmentation performance.

To address the above challenges, previous attempts at polyp segmentation employed conventional methods, including region growing~\cite{Gonzalez2018region}, deformable models~\cite{Yao2004deformable}, fuzzy C-means clustering methods~\cite{Yao2004deformable} and threshold-based methods~\cite{Guo2021threshold}. However, the reliance on handcrafted features limited the ability of these methods to capture complex representations, resulting in suboptimal segmentation results~\cite{Gupta2024polreview}. To mitigate these limitations, researchers have increasingly turned to encoder-decoder architectures as a promising avenue for polyp segmentation. Existing medical image segmentation research~\cite{Dumitru2023ducknet, Haider2023mfranet} emphasizes the importance of sharing contextual information between the encoder and decoder, given the target object's unique shapes, sizes, and textures. For instance, Li-SegPNet~\cite{Sharma2023LiSegPNet} leverages cross-dimensional feature interaction within the encoder block. Other approaches such as FCBFormer~\cite{Sanderson2022fcbformer}, MSRF-Net~\cite{Srivastava2022msrfnet}, and CRCNet~\cite{Zhu2023crcnet} incorporate multi-scale feature learning to capture features at various scales. Additionally, some studies enhance convolutional network capabilities by incorporating attention mechanisms. For example, PraNet~\cite{Fan2020pranet}, EGTransUNet~\cite{Pan2023egtransunet}, TGANet~\cite{Tomar2022tganet} and CFHA-Net~\cite{Yang2023cfhanet}.

Despite these advances, capturing the full spectrum of complex polyp variations using solely convolutional-based feature extraction remains challenging. Leveraging transformer networks can offer a viable solution for extracting global features. However, the irregular nature of polyps poses a significant obstacle for transformer-based approaches, hindering their ability to locate and extract these features accurately. Therefore, developing a novel approach that effectively unlocks the potential of spatial-global features integrated with multi-scale feature extraction could revolutionize the segmentation of variable polyp shapes. Motivated by this, we propose a framework that captures integrated multi-scale spatial features with spatial aware global dependencies while prioritizing the boundary information. This integrated approach aims to achieve superior accuracy and robustness.

We address the limitations of prior approaches in handling integrated spatial-global dependencies with multi-scale features by introducing a novel architecture, A Multiscale Network with Spatial-enhanced Attention (MNet-SAt), for polyp segmentation. MNet-SAt utilizes an encoder-decoder framework enhanced with edge-guided feature refinement to preserve polyp boundary information. Furthermore, MNet-SAt leverages a synergistic architecture featuring multi-scale aggregation coupled with a spatial enhanced attention mechanism, enabling it to learn detailed local features at different scales effectively and spatially enhanced global patterns. We demonstrate the efficacy of MNet-SAt by achieving state-of-the-art results on two benchmark datasets: Kvasir-SEG~\cite{Jha2020kvasir} and CVC-ClinicDB~\cite{Bernal2015cvc}. The following key findings highlight the significant contributions of this study:

\begin{itemize}
    \item We propose MNet-SAt, a Multiscale Network with Spatial-enhanced Attention, for the segmentation of polyps in colonoscopy. This framework leverages an Encoder-Decoder architecture composed of Edge-Guided Feature Enrichment (EGFE) units to capture fine-grained information and edge features at various levels, facilitating the accurate learning of polyp boundaries.
    \item We design a Multi-Scale Feature Aggregator (MSFA) unit to extract diverse contextual features of polyps, leveraging variable receptive fields across the channel's spatial dimension to improve predictive performance.
    \item We introduce a Spatial-Enhanced Attention (SEAt) unit to learn global contextual features and establish robust relationships among multi-scale aggregated features, effectively highlighting the region of interest.
    \item We integrate Channel-Enhanced Atrous Spatial Pyramid Pooling (CE-ASPP), utilizing multiple levels of atrous convolution to learn diverse multi-scale features and fusing with inter-channel information.
    \item MNet-SAt outperforms ten state-of-the-art polyp segmentation methods on two benchmark datasets, demonstrating superior performance in segmenting polyps with diverse and challenging morphologies, highlighting its potential for clinical translation. 
\end{itemize}


\section{Related Works}
\subsection{Encoder Decoder-based Segmentation}
Deep learning, notably with Convolutional Neural Networks (CNNs), has transformed medical imaging and radiology, excelling in object detection, segmentation tasks, and image classification~\cite{Litjens2017dl, Minaee2022segdl}. The U-Net architecture~\cite{Ronneberger2015unet} stands as a pivotal development in this field. Characterized by an encoder-decoder CNN design with strategically placed multi-scale skip connections, U-Net facilitates the preservation of high-fidelity image information, which is crucial for the precise localization of anatomical structures and pathological regions in medical images. Building on the U-Net architecture, researchers have proposed numerous segmentation models, including UNet++~\cite{Zhou2018unet++}, modified-UNet~\cite{Seo2020munet}, ResUNet~\cite{Diakogiannis2020resunet}, and ResUNet++~\cite{Jha2019resunet++}.

Dumitru et al.~\cite{Dumitru2023ducknet} introduced DUCK-Net, a Deep Understanding of Convolutional Kernal Network, which employs a residual downsampling mechanism and a specialized convolutional block to extract multi-resolution image information within its encoder effectively. Haider et al.~\cite{Haider2023mfranet} introduced the Colorectal Cancer Segmentation Network (CCS-Net) and a multi-scale network that effectively retains and aggregates features across various scales to capture both lower-level and high-level information through feature fusion precisely. Sharma et al.~\cite{Sharma2023LiSegPNet} introduced Li-SegPNet, leveraging cross-dimensional interaction in feature maps with a pre-trained ResNet50 backbone~\cite{He2016deeprl} and a novel encoder block incorporating modified triplet attention.

\subsection{Multi-Scale Feature-based Segmentation}
Accurate segmentation maps preserve high-resolution representations within the segmentation architecture, ensuring the capture of fine spatial details~\cite{Wang2021deep}. Multi-scale feature fusion emerges as an alternative to up-sampling lower-resolution representations. Wang et al.~\cite{Wang2021deep} demonstrated that exchanging features across scales facilitates the propagation of high-resolution information, leading to accurate segmentation maps. To achieve this, we perform parallel processing of features at different scales, followed by their effective combination.

Sanderson et al.~\cite{Sanderson2022fcbformer} introduced a novel architecture that combines a feature-extraction module using a Fully Convolutional Network (FCN) with a Transformer. The extracted features are then fused to produce the final segmentation map. Srivastava et al.~\cite{Srivastava2022msrfnet} introduced a residual-based multiscale fusion network that utilizes dual-scale blocks for sequential exchange and fusion of multi-scale features. Zhu et al.~\cite{Zhu2023crcnet} present CRCNet, a dual-encoder decoder segmentation framework incorporating a Global-Local Context Module (GLCM) for efficient multi-scale feature extraction and a Multi-Modality Cross Attention (MMCA) module to achieve selective feature fusion.

\subsection{Attention-based Segmentation}
Hu et al.~\cite{Hu2020se} introduced channel-wise attention with Squeeze and Excitation Network (SE-Net). The SE-Net block models inter-dependencies between channels, extracting a global information map that focuses on primary features and suppresses irrelevant ones. Su et al.~\cite{Su2023fednet} modify the U-Net architecture by incorporating Squeeze-and-Excitation (SE) blocks. Kaul et al.~\cite{Kaul2019focusnet} proposed FocusNet, a dual-branch parallel network architecture that leverages both channel-wise and spatial mechanisms for feature extraction and representation. Fan et al.~\cite{Fan2020pranet} introduced PraNet, a segmentation network employing a Parallel Partial Decoder (PPD) for high-level feature compression and an attention module for refined boundary identification. Pan et al.~\cite{Pan2023egtransunet} integrated a Transformer architecture with attention mechanisms into the encoder-decoder pair of their network to improve semantic localization and feature discriminancy.

Tomar et al.~\cite{Tomar2022tganet} proposed a polyp segmentation system that leverages size and polyp number information through text attention. This approach, achieved via an auxiliary classification task, facilitates the network's learning of distinct feature representations for various polyp sizes and its adaptation to multi-polyp scenarios. Yang et al.~\cite{Yang2023cfhanet} proposed a feature fusion strategy that includes cross-scaling of features and a Triple Hybrid Attention (THA) mechanism. THA bridges the encoder-decoder pair, combining channel and spatial attention to capture long-range dependencies within polyp features.



\section{Methodology}
\subsection{Overview of Framework}
\begin{figure*}[!ht]
  \centering
  \includegraphics[width=\textwidth]{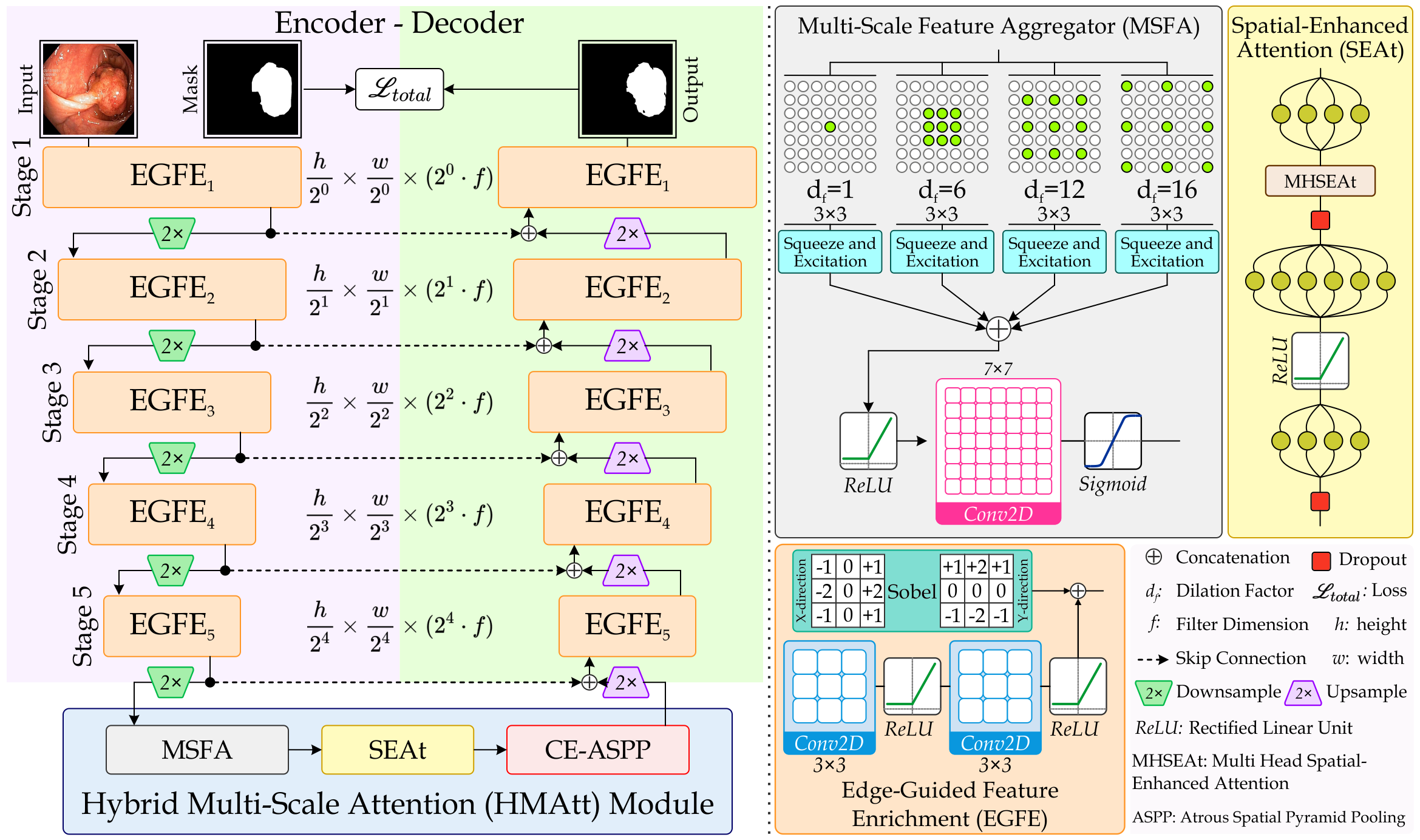}
  \caption{MNet-SAt framework for polyp segmentation employs a U-shaped architecture with an encoder and decoder linked by skip connections. Edge-Guided Feature Enrichment (EGFE) units within each stage enable feature transfer by extracting features from polyp anatomy, enhancing the segmentation task. A Hybrid Multi-Scale Attention (HMAtt) module bridges the encoder and decoder, learning multi-scale features by prioritizing important regions.}
  \label{fig:mnetsat}
\end{figure*}

The MNet-SAt framework illustrated in~\cref{fig:mnetsat} adopts a U-shaped architecture with an encoder and decoder pathway linked through skip connections. The encoder downsamples input data of dimensions $h \times w$ using convolutional layers, while the decoder restores the data to its original dimensions. Each encoder stage $i$ $(i \in [1,2 \ldots 5])$ downscales features to dimensions of $\frac{h}{2^{i-1}} \times \frac{w}{2^{i-1}}$. Key components of this framework include the Edge-Guided Feature Enrichment (EGFE) unit and the Hybrid Multi-Scale Attention (HMAtt) module. The EGFE unit integrates edge-aware feature extraction and enhances boundary-aware feature representation, improving segmentation performance. The HMAtt module employs multi-scale receptive fields to enrich the extracted features, enhancing the framework's adaptability. Additionally, the framework achieves precise anomaly localization and segmentation by focusing on critical regions. Finally, a convolutional layer with an appropriate activation function processes the aggregated features from the decoder's last stage to compute segmentation probabilities.

\subsection{Encoder-Decoder Backbone}
\label{sec:edb}
Our MNet-SAt framework utilizes a U-shaped backbone structured with an encoder and a decoder, each containing five Edge-Guided Feature Enrichment (EGFE) units designed to generate feature maps rich in edge information. To incorporate boundary cues, we apply a Sobel operator before the convolution within each EGFE unit. This enables the separate learning of coarse boundaries and effective edge features. In the encoder path, a downsampling operation (reducing resolution by $2 \times 2$) precedes each stage to progressively reduce the feature map size and feed it into the next encoder. The decoding path commences from the output of the HMAtt module, a multi-scale attention network that facilitates communication between the encoder-decoder pair. The decoder pathway in MNet-SAt leverages encoder features provided by the EGFE units to generate segmentation masks. Decoding blocks upsample their outputs by $2 \times 2$ before concatenation with corresponding encoder feature maps. We then feed the concatenated features from multiple stages into subsequent decoding blocks for further refinement. Finally, we upsample the last stage's feature map to match the input dimension, incorporating information from the encoder and decoder. \autoref{table:fmap} details the feature map dimensions for each stage in the encoder and decoder.

\begin{table}[!ht] \small
\caption{Feature Map Dimensions in MNet-SAt Stages.}
     \begin{tabularx}{\linewidth}{@{}l *4{C} @{}}
        \toprule                
        \textbf{Stage}&\textbf{Encoder} &\textbf{Output} &\textbf{Decoder} &\textbf{Output} \\
        &\textbf{Blocks} &\textbf{Dimensions} &\textbf{Blocks} &\textbf{Dimensions} \\
    \midrule
        Stage 1 & $\text{EGFE}_1$ & $512 \times 512 \times 64$ & $\text{EGFE}_1$ & $512 \times 512 \times 3$ \\
        & Down & $256 \times 256 \times 64$ & Up & $512 \times 512 \times 64$ \\
    \midrule
        Stage 2 & $\text{EGFE}_2$ & $256 \times 256 \times 128$ & $\text{EGFE}_2$ & $256 \times 256 \times 64$ \\
        & Down & $128 \times 128 \times 128$ & Up & $256 \times 256 \times 128$ \\
    \midrule
        Stage 3 & $\text{EGFE}_3$ & $128 \times 128 \times 256$ & $\text{EGFE}_3$ & $128 \times 128 \times 128$ \\
        & Down & $64 \times 64 \times 256$ & Up & $128 \times 128 \times 256$ \\
    \midrule
        Stage 4 & $\text{EGFE}_4$ & $64 \times 64 \times 512$ & $\text{EGFE}_4$ & $64 \times 64 \times 256$ \\
        & Down & $32 \times 32 \times 512$ & Up & $64 \times 64 \times 512$ \\
    \midrule
        Stage 5 & $\text{EGFE}_5$ & $32 \times 32 \times 1024$ & $\text{EGFE}_5$ & $32 \times 32 \times 512$ \\
        & Down & $16 \times 16 \times 1024$ & Up & $32 \times 32 \times 1024$ \\
    \bottomrule        
    \end{tabularx}
\label{table:fmap}
\end{table}

The MNet-SAt leverages skip connections between corresponding EGFE units. These connections mitigate information loss during encoder downsampling by bridging the gap between low-resolution and high-resolution feature maps. Conversely, during decoder upsampling, they provide rich contextual information to higher-resolution layers, facilitating the generation of finer details in the final segmentation mask.

\subsection{Edge-Guided Feature Enrichment Unit}
\label{sec:egfe}
The scarcity of annotated data in clinical practice hinders existing methods' ability to capture polyp boundaries accurately. To address this, we propose the Edge-Guided Feature Enrichment (EGFE) unit, designed to enhance boundary quality in segmentation masks. EGFE robustly preserves edge information, mitigating the issue of weak boundaries, bridges the semantic gap between low-level and high-level features, and directly captures spatial features of polyps from input images. A pre-processing step incorporating a 2D Sobel operator~\cite{Wang2024sbcnet} ensures the preservation of crucial edge details during convolutions.

\begin{figure}[!b]
\captionsetup[subfigure]{justification=centering}
     \centering
     \begin{subfigure}[t]{0.2\linewidth}\centering
         \includegraphics[width=0.69\linewidth]{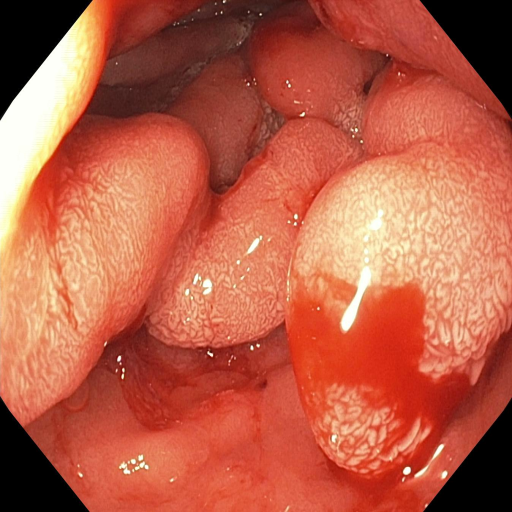}
         \caption{Polyp 1}\label{fig:ks1}
     \end{subfigure}\hfill
     \begin{subfigure}[t]{0.2\linewidth}\centering
         \includegraphics[width=0.69\linewidth]{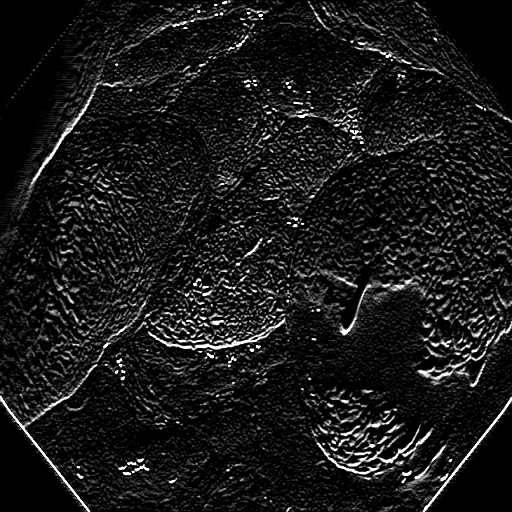}
         \caption{Sobel x}\label{fig:ksx}
     \end{subfigure}\hfill
     \begin{subfigure}[t]{0.2\linewidth}\centering
         \includegraphics[width=0.69\linewidth]{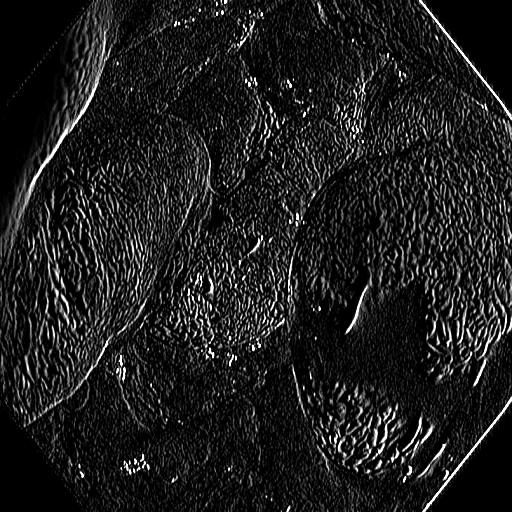}
         \caption{Sobel y}\label{fig:ksy}
     \end{subfigure}\hfill
     \begin{subfigure}[t]{0.2\linewidth}\centering
         \includegraphics[width=0.69\linewidth]{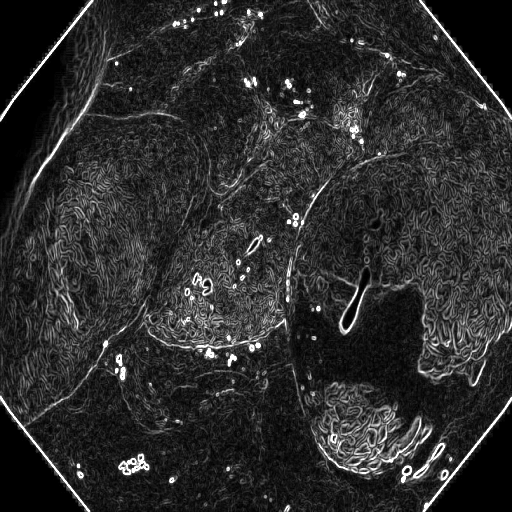}
         \caption{Sobel xy}\label{fig:kse}
     \end{subfigure}
     
     \begin{subfigure}[t]{0.2\linewidth}\centering
         \includegraphics[width=0.69\linewidth]{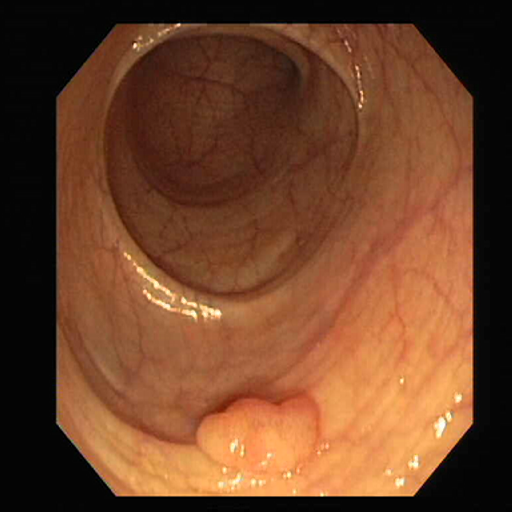}
         \caption{Polyp 2}\label{fig:cd1}
     \end{subfigure}\hfill
     \begin{subfigure}[t]{0.2\linewidth}\centering
         \includegraphics[width=0.69\linewidth]{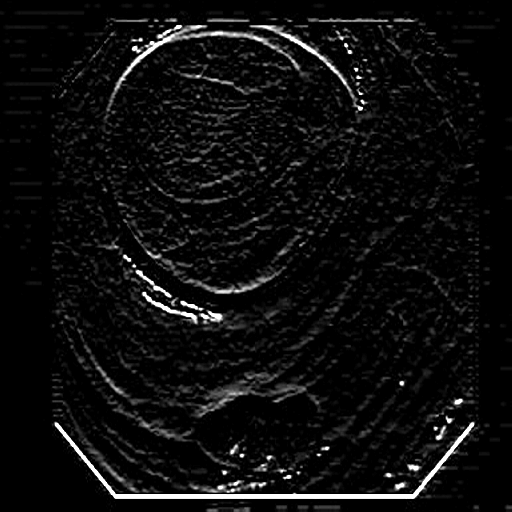}
         \caption{Sobel x}\label{fig:cdx}
     \end{subfigure}\hfill    
     \begin{subfigure}[t]{0.2\linewidth}\centering
         \includegraphics[width=0.69\linewidth]{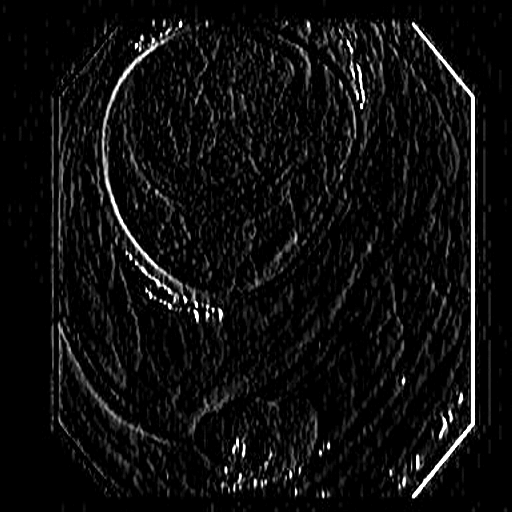}
         \caption{Sobel y}\label{fig:cdy}
     \end{subfigure}\hfill
     \begin{subfigure}[t]{0.2\linewidth}\centering
         \includegraphics[width=0.69\linewidth]{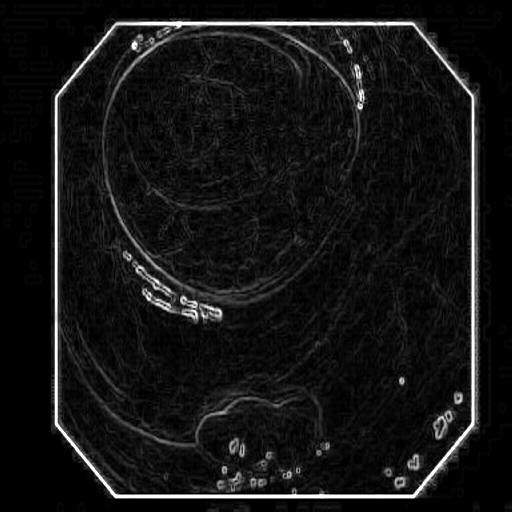}
         \caption{Sobel xy}\label{fig:cde}
     \end{subfigure}
    \caption{Edge maps of the Sobel operator in x-y directions}
    \label{fig:sobelxy}
\end{figure}

The Edge-Guided Feature Enrichment (EGFE) unit actively enhances feature representation by integrating edge knowledge into the set of features. It applies a Sobel operator to the input feature set, generating separate gradient magnitude maps for horizontal $(\mathcal{S}_x)$ and vertical $(\mathcal{S}_y)$ directions (\cref{fig:sobelxy}). As illustrated in~\cref{fig:sobelxy}, the Sobel operation in a single direction captures limited information about the polyp. However, combining these maps (\cref{fig:kse} and~\cref{fig:cde}) effectively captures polyp boundaries and enriches learned features. The final gradient magnitude map, produced using the Euclidean norm (\cref{eq:sobel}), highlights regions with significant intensity changes at polyp boundaries.

\begin{align}
\label{eq:sobel}
\mathcal{S} = \sqrt{\mathcal{S}_x^2 + \mathcal{S}_y^2}
\end{align}

\noindent Here, the output of the prior edge detector is denoted by $\mathcal{S}$, representing the gradient magnitude map. This map is then element-wise added to the output of the convolution layers for feature fusion. Each EGFE unit consists of two convolutional layers with $f$ filters of kernel size $3 \times 3$ and stride $2$. The number of filters, $f$, progressively increases at each stage according to the formula $64 \cdot 2^{i-1}$, where $i$ represents the stage number. ReLU activation functions are applied after each convolution to induce non-linearity. This allows the framework to learn and extract complex, hidden relationships within the data. \cref{eq:egfeunit} formalizes this process:

\begin{align}
\label{eq:egfeunit}
\mathcal{F}_{egfe}^i = \mathcal{S} \otimes \biggl( \underset{ReLU}{\alpha} \Bigl(Conv_2\bigl(\underset{ReLU}{\alpha}(Conv_1 (\mathcal{F}_{egfe}^{i-1}))\bigr)\Bigr)\biggr)
\end{align}

\noindent where $\mathcal{S}$, $\underset{ReLU}{\alpha}$, and $Conv_j$ denotes 2D Sobel operation, ReLU activation function, the number of 2D convolutions with $j=\{1,2\}$ respectively, and $\mathcal{F}_{egfe}^i$ represents the edge-guided features learned at the $i$-th layer. This element-wise addition facilitates the fusion of edge information extracted by the Sobel operator with the spatial features learned by the convolutional layers. This fusion enables the framework to focus on and refine the representation of polyps, ultimately contributing to enhanced segmentation performance.

\subsection{Hybrid Multi-Scale Attention Module}
The MNet-SAt framework utilizes a Hybrid Multi-Scale Attention (HMAtt) module to bridge encoder-decoder backbone. HMAtt effectively captures multi-scale contextual information and highlights salient regions within feature maps.

\subsubsection{Multi-Scale Feature Aggregator Unit}
Colonoscopy image analysis often assumes that specific features reside at specific scales~\cite{Zheng2024cgmanet}. However, existing multi-scale approaches primarily focus on feature extraction~\cite{Zheng2024cgmanet}, neglecting aggerated channel-spatial dimensions. To mitigate this, we introduce a Multi-Scale Feature Aggregator (MSFA) Unit that integrates Squeeze-and-Excitation (SE) blocks~\cite{Hu2020se} after multi-scale feature extraction and aggregates them. This emphasis on salient regions enhances the polyp's feature representation.

The MSFA unit receives the output of the encoder's final EGFE unit, $\mathcal{F}_{egfe}^5$, as input. It then processes this feature map through multiple convolutions with varying dilation rates, as defined mathematically in~\cref{eq:dilconv}.

\begin{align}
\label{eq:dilconv}
\mathcal{F}_{k,d_r}[l]=\sum_{\kappa=1}^k \mathcal{F}_{egfe}^5[l+d_r \ast \kappa] \ast \varphi_{\kappa}
\end{align}

\noindent Here, we define $l$ as the location on the feature map, $d_r \in \{1,6,12,18\}$ as the dilation rate, $\varphi_{\kappa}$ as the $\kappa-$th parameter of convolution filter, and $k$ signifies the filter size. \cref{eq:dilconv} illustrates varying dilation rates $(d_r)$ enables the acquisition of diverse receptive fields,  serving as multiple scales to learn spatial components. Afterward, we recalibrate these multi-scale features through the SE block to prioritize the relevant multi-scale features. By concatenating these multi-scale aggregated features with distinct receptive fields, MSFA generates high-level contextual features. The following equations represent the mathematical formulation of MSFA.

\begin{subequations}
    \begin{equation}
    \label{eq:msconcat}
        \mathcal{F}_{\oplus} =Concat \Bigl(SE \bigl(\mathcal{F}_{3,i} (\mathcal{F}_{egfe}^5) \bigr)\Bigr), i \in (1,6,12,18)
    \end{equation}
        
    \begin{equation}
    \label{eq:msfeat}
        \mathcal{F}_{msfa} = \underset{softmax}{\alpha} \Bigl(\mathcal{F}_{7,1}\bigl(\underset{ReLU}{\alpha}(\mathcal{F}_{\oplus})\bigr)\Bigr)
    \end{equation}
\end{subequations}

\noindent Here, $\mathcal{F}_{3, i}$ represents the multi-scale features captured at different dilation rates $(1,6,12,18)$ enhanced by SE block, $\mathcal{F}_{\oplus}$ denotes the concatenated multi-scale features which are further processed by $\underset{ReLU}{\alpha}$, ReLU activation function. Finally, all concatenated features are processed through convolution with $7 \times 7$ kernel to generate final multi-scale aggregated features, denoted by $\mathcal{F}_{msfa}$.

\subsubsection{Spatial-Enhanced Attention Unit}
As illustrated in~\cref{fig:mnetsat}, the Spatial-Enhanced Attention (SEAt) unit, a core component of the HMAtt module, captures spatial-aware global dependencies within the multi-scale aggregated features $\mathcal{F}_{msfa}$ derived from the MSFA unit. The SEAt comprises a Fully Connected (FC) layer, MHSEAt, feed-forward layers, and a dropout layer. Due to the pivotal role of the attention mechanism within SEAt, we provide a detailed introduction to MHSEAt below.

\begin{figure}
  \centering
  \includegraphics[width=0.75\linewidth]{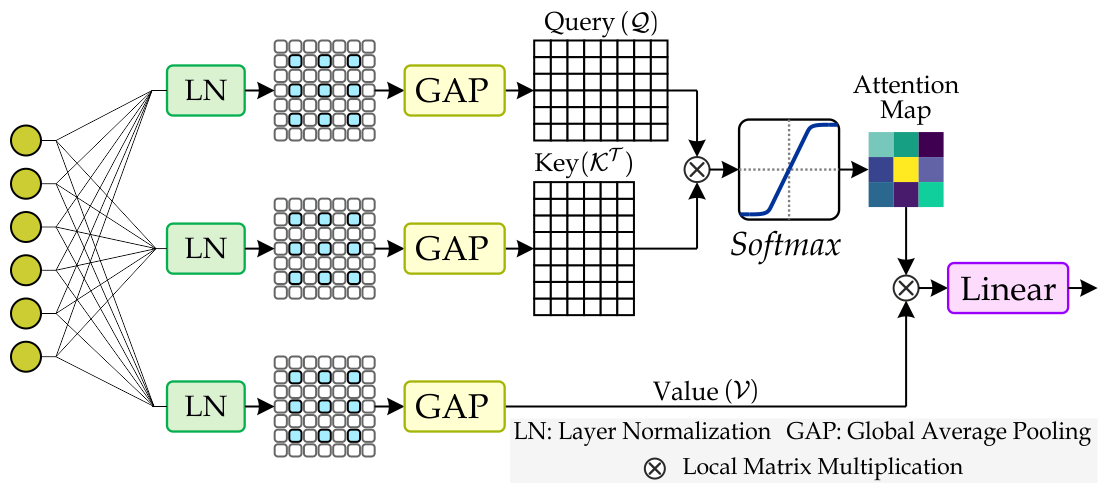}
  \caption{Multi-Head Spatial-Enhanced Attention}
  \label{fig:mhseat}
\end{figure}

\textbf{Multi-Head Spatial-Enhanced Attention (MHSEAt):} The Multi-Head Spatial-Enhanced Attention (MHSEAt) performs the self-attention operation along the spatial dimension. As shown in~\cref{fig:mhseat}, firstly, the three identical copies of feature maps are normalized through Layer Normalization $(LN)$, followed by convolution operation $(Conv)$ with $3 \times 3$ kernel having dilation rate of $2$. Next, the spatial-enhanced features are aggregated with Global Average Pooling $(GAP)$ to obtain the Query ($\mathcal{Q}$), Key ($\mathcal{K}$), and Values ($\mathcal{V}$). We can formulate the process mentioned above as follows:

\begin{subequations}
\begin{equation}
    \mathcal{Q} = LN (\underset{3 \times 3}{Conv} (GAP (X)))W_{\mathcal{Q}}
\end{equation}

\begin{equation}
    \mathcal{K} = LN (\underset{3 \times 3}{Conv} (GAP (X)))W_{\mathcal{K}}
\end{equation}

\begin{equation}
\label{eq:qkv}
    \mathcal{V} = LN (\underset{3 \times 3}{Conv} (GAP (X)))W_{\mathcal{V}}
\end{equation}
\end{subequations}

\begin{equation}
\label{eq:seattention}
    X_{SEAt} = \underset{softmax}{\alpha} \left( \frac{\mathcal{Q} \cdot \mathcal{K}^T}{\sqrt{d_{\mathcal{K}}}} \right) \cdot \mathcal{V}
\end{equation}

\noindent Here, $X$ is the input feature map, $W_{\mathcal{Q}}, W_{\mathcal{K}}, W_{\mathcal{V}} \in \mathbb{R}^{c \times c}$ are the projection matrices, $\underset{softmax}{\alpha}$ denotes the softmax function, $\sqrt{d_{\mathcal{K}}}$ denotes the dimension of the key and $X_{SEAt}$ is the feature map after linear spatial reduction multi-head spatial-enhanced attention.

The SEAt unit follows the MHSEAt layer with two stacked Fully Connected (FC) layers. Each layer uses the ReLU activation (denoted by $\underset{ReLU}{\alpha}$) to incorporate non-linearity, allowing the network to model complex data relationships. Formally, $\mathcal{F}_{SEAt} = FC_2 \Bigl(\underset{ReLU}{\sigma}\bigl( FC_1 (\mathcal{F}_{MHSEAt}) \bigr)\Bigr)$, where $\mathcal{F}_{MHSEAt}$ represents the spatial-enhanced attention features. This process effectively captures local and global dependencies while preserving spatial information in the final output, $\mathcal{F}_{SEAt}$.

\subsubsection{Channel-Enhanced Atrous Spatial Pyramid Pooling}
To improve polyp representation by resampling features at different scales, we propose Channel-Enhanced Atrous Spatial Pyramid Pooling (CE-ASPP). Unlike the standard Atrous Spatial Pyramid Pooling (ASPP)~\cite{Chen2018aspp}, which uses dilated convolutions for multiscale extraction and standard convolutions for fusion, CE-ASPP incorporates a three-step refinement to recalibrate the importance of various feature channels adaptively.

\begin{figure}[!ht]
  \centering
  \includegraphics[width=0.75\linewidth]{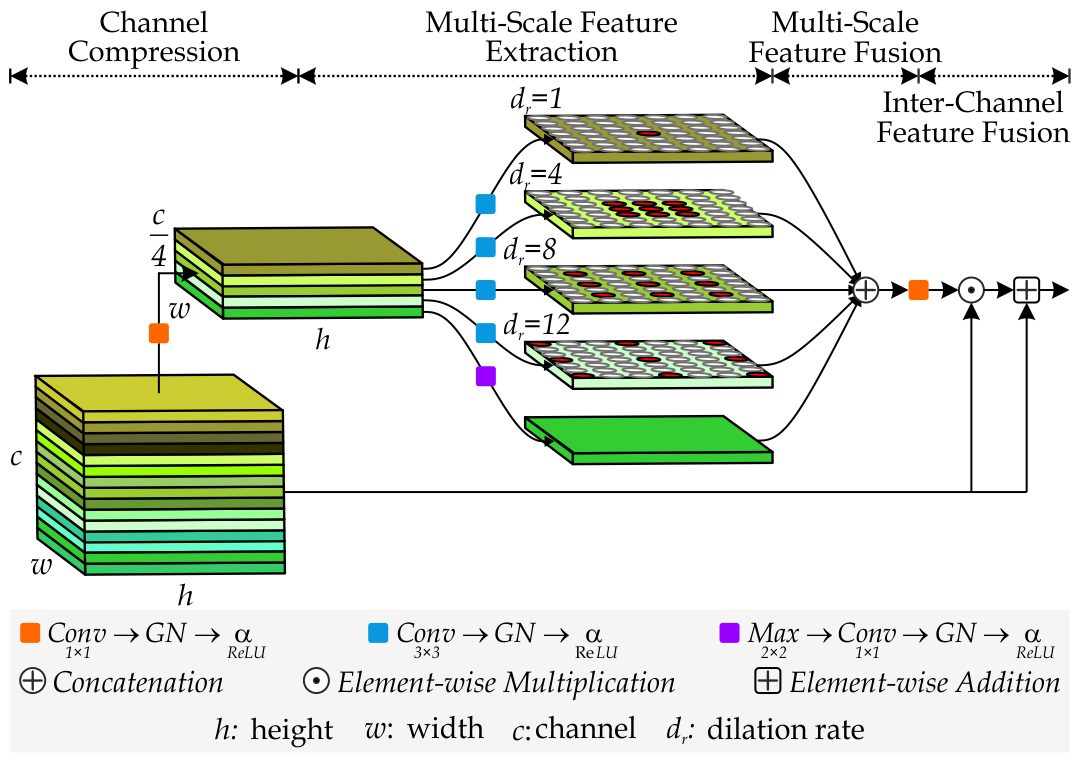}
  \caption{Channel-Enhanced Atrous Spatial Pyramid Pooling}
  \label{fig:ceaspp}
\end{figure}
 
Within CE-ASPP (refer to~\cref{fig:ceaspp}), we first compress the input feature channels by a factor of four. This compression is then reversed before the final output is generated (refer~\cref{eq:ce1}).

\begin{equation}
\label{eq:ce1}
    \mathcal{F}_c = \underset{ReLU}{\alpha} \Bigl(GN \bigl( \underset{1 \times 1}{Conv} (\mathcal{F}_{SEAt}) \bigr)  \Bigr)
\end{equation}

\noindent Here, $\underset{ReLU}{\alpha}$ denotes the ReLU activation function, $\mathcal{F}_c$ is the channel-compressed features, and GN is Group Normalization~\cite{Wu2020gn}, chosen for its accuracy stability across batch sizes.

Secondly, $\mathcal{F}_c$ undergoes five parallel routing convolutions: four with $3 \times 3$ depth-wise atrous convolutions (dilation rates 1, 4, 8, and 12) and one with max-pooling followed by $1 \times 1$ point-wise convolution for salient feature extraction and channel dimension adjustment. This process is expressed below:

\begin{subequations}
    \begin{align}    
        \mathcal{F}_c^{d_r} = \underset{ReLU}{\alpha} \Bigl(GN \bigl( \underset{3 \times 3}{Conv} (\mathcal{F}_c) \bigr)  \Bigr), d_r \in [1,4,8,12]
    \end{align}
    \begin{align}    
    \label{eq:ce2}
        \mathcal{F}_c^{Max} = \underset{ReLU}{\alpha} \Bigl(GN \bigl( \underset{1 \times 1}{Conv} (\underset{2 \times 2}{Max}(\mathcal{F}_c)) \bigr)  \Bigr)
    \end{align}
\end{subequations}

\noindent where, $\mathcal{F}_c^{d_r}$ and $\mathcal{F}_c^{Max}$ denote the compressed multi-scale features obtained from atrous convolutions (varying dilation rates) and max-pooling, respectively. Lastly, Multi-scale and inter-channel fusion are then applied to generate $\mathcal{F}_{ce-aspp}$ as expressed in~\cref{eq:ce3a} and~\cref{eq:ce3b}.

\begin{equation}
\label{eq:ce3a}
    \mathcal{F}_{\oplus} = Concat(\mathcal{F}_c^{1}, \mathcal{F}_c^{4}, \mathcal{F}_c^{8}, \mathcal{F}_c^{12}, \mathcal{F}_c^{Max})
\end{equation}
\begin{equation}
\label{eq:ce3b}
    \mathcal{F}_{ce-aspp} =  \mathcal{F}_{SEAt} \odot \underset{ReLU}{\alpha} \Bigl(GN \bigl( \underset{1 \times 1}{Conv} (\mathcal{F}_{\oplus}) \bigr)  \Bigr) \boxplus \mathcal{F}_{SEAt}
\end{equation}

\noindent Here, $\mathcal{F}_{\oplus}$ denotes the concatenated multi-scale features. $\boxplus$ and $\odot$ represent element-wise multiplication and addition, respectively. The CE-ASPP module generates the final feature, $\mathcal{F}_{ce-aspp}$ is fed into the decoder.

\subsection{Loss Computation}
Our framework tackles class imbalance common in medical segmentation using a total loss, $\mathscr{L}_{total}$, that incorporates Dice loss $(\mathscr{L}_{dice})$ and binary cross-entropy loss $(\mathscr{L}_{bce})$. The total loss function is defined as~\cref{eq:loss}. Dice loss is further used for parameter optimization, maximizing Dice score and IoU.

\begin{subequations}
\begin{equation}
    \mathscr{L}_{Dice} = 1 - \frac{2 \sum (y_{g} \cap y_{p}) + \epsilon}{\sum(y_{g} + y_{p}) + \epsilon}
\end{equation}
\begin{equation}
    \mathscr{L}_{BCE} = -[y_{g} \cdot \log(y_{p})+(1-y_{g})\cdot \log(1-y_{p})]
\end{equation}
\end{subequations}
\begin{equation}
    \label{eq:loss}
    \mathscr{L}_{total} = \gamma\mathscr{L}_{Dice} + \delta\mathscr{L}_{BCE}
\end{equation}

\noindent Here, $y_{g}$ and $y_{p}$ represent the ground-truth and predicted labels, respectively, and $\epsilon$ is a smoothing factor. For balanced contributions from both loss terms, we set $\gamma$ and $\delta$ to 0.5.


\section{Experimental Study and Results}
\subsection{Benchmark Datasets}
\textit{Kvasir-SEG}~\cite{Jha2020kvasir}:
Provided by Vestre Viken Health Trust (VV), Norway, it consists of 1000 gastrointestinal polyp images collected with endoscopic equipment and meticulously labeled by experienced gastroenterologists. Image resolutions range between $332 \times 487$ pixels to $1920 \times 1072$ pixels.

\textit{CVC-ClinicDB}~\cite{Bernal2015cvc}:
A widely used biomedical dataset comprises 612 images ($384 \times 288$ pixels)  extracted from 31 colonoscopy videos collected at the Hospital Clinic in Barcelona, Spain. This dataset served as training data for Automatic Polyp Detection in the MICCAI 2015 Sub-Challenge.

\subsection{Experimental Setup}
\subsubsection{Implementation Details}
We conducted our experiments using Python and the TensorFlow library on an NVIDIA P100 GPU with 16GB GPU RAM and 512GB system RAM. We optimized model training using the Adam optimizer with an initial learning rate of 1e-4 and a Keras callback to reduce the learning rate upon plateauing accuracy. We augmented the training datasets by applying random translations, rotations, contrast adjustments, and mirroring, generating eight augmented images for each original image. We then split the data into an 8:1:1 ratio for training, validation, and testing. All images were resized to $512 \times 512$ pixels, and we used a batch size of 8 to manage memory requirements and enhance training stability. We trained the MNet-SAt framework for 70 epochs, a balance chosen to achieve convergence while mitigating overfitting. We maintained consistent experimental settings across all evaluations to ensure a fair comparison.

\subsubsection{Evaluation Metrics}
We evaluated performance using four metrics: Dice Similarity Coefficient (DSC), Intersection over Union (IoU), Precision (Pre), and Recall (Rec), quantifying the similarity between the predicted map and ground-truth~\cite{Zheng2024cgmanet}.

\subsection{Experimental Result Analysis}
\subsubsection{Comparisons With Existing Baselines}
To comprehensively evaluate the effectiveness of MNet-SAt, we compare its performance against several recent deep learning-based segmentation methods. These include three encoder-decoder networks: DUCK-Net~\cite{Dumitru2023ducknet}, MFRA-Net~\cite{Haider2023mfranet}, Li-SegPNet~\cite{Sharma2023LiSegPNet}, three multi-scale feature-based methods: FCBFormer~\cite{Sanderson2022fcbformer}, MSRF-Net~\cite{Srivastava2022msrfnet}, CRCNet~\cite{Zhu2023crcnet}, and four attention-based segmentation approaches: PraNet~\cite{Fan2020pranet}, EGTransUNet~\cite{Pan2023egtransunet}, TGANet~\cite{Tomar2022tganet}, CFHA-Net~\cite{Yang2023cfhanet}.

\begin{table*}[!b] \small
\caption{Quantitative comparison of the proposed MNet-SAt with existing baselines on Kvasir-SEG and CVC-ClinicDB employing 5-fold cross-validation. Higher values for DSC, IoU, Pre, and Rec indicate better performance. \textcolor{red}{\textbf{Red}} and \textcolor{ib}{\textbf{{blue}}} represent the best and second-best results, with $|\nabla|$ denoting the absolute performance drop compared to MNet-SAt.}
    \begin{tabularx}{\linewidth}{@{}L{2.75cm} *4{C} *4{C} @{}}    
        \toprule

        \textbf{Methods}
        &\multicolumn{4}{c}{\textbf{Kvasir-SEG}} 
        &\multicolumn{4}{c}{\textbf{CVC-ClinicDB}} \\
        \cmidrule(l){2-5} \cmidrule(l){6-9} 
        
        &\textbf{DSC$_{|\nabla|}$} &\textbf{IoU$_{|\nabla|}$} &\textbf{Pre$_{|\nabla|}$} &\textbf{Rec$_{|\nabla|}$}        
        &\textbf{DSC$_{|\nabla|}$} &\textbf{IoU$_{|\nabla|}$} &\textbf{Pre$_{|\nabla|}$} &\textbf{Rec$_{|\nabla|}$} \\
        
        \midrule    
        \multicolumn{9}{l}{\textbf{Encoder Decoder-based Segmentation}} \\
        \midrule
        
        DUCK-Net~\cite{Dumitru2023ducknet}
        & \textcolor{ib}{\textbf{95.56}}\scriptsize$_{\textit{1.05}}$ 
        & 90.53\scriptsize$_{\textit{6.39}}$  
        & 95.32\scriptsize$_{\textit{2.04}}$  
        & 93.45\scriptsize$_{\textit{3.67}}$ 
        & \textcolor{ib}{\textbf{97.15}}\scriptsize$_{\textit{1.45}}$  
        & 88.55\scriptsize$_{\textit{6.99}}$ 
        & 94.02\scriptsize$_{\textit{2.87}}$ 
        & 96.39\scriptsize$_{\textit{0.26}}$  \\
        
        MFRA-Net~\cite{Haider2023mfranet}
        & 94.37\scriptsize$_{\textit{2.24}}$ 
        & \textcolor{ib}{\textbf{90.76}}\scriptsize$_{\textit{6.16}}$  
        & 92.28\scriptsize$_{\textit{5.08}}$  
        & \textcolor{ib}{\textbf{96.69}}\scriptsize$_{\textit{0.43}}$ 
        & 96.24\scriptsize$_{\textit{2.36}}$  
        & 93.36\scriptsize$_{\textit{2.18}}$ 
        & 96.83\scriptsize$_{\textit{0.06}}$ 
        & 94.37\scriptsize$_{\textit{1.76}}$  \\
        
        Li-SegPNet~\cite{Sharma2023LiSegPNet}
        & 91.86\scriptsize$_{\textit{4.75}}$ 
        & 89.17\scriptsize$_{\textit{7.75}}$  
        & 92.79\scriptsize$_{\textit{4.57}}$  
        & 87.88\scriptsize$_{\textit{9.24}}$ 
        & 91.85\scriptsize$_{\textit{6.75}}$  
        & 93.55\scriptsize$_{\textit{1.99}}$ 
        & \textcolor{red}{\textbf{96.99}}\scriptsize$_{\textit{0.10}}$ 
        & 87.09\scriptsize$_{\textit{9.04}}$ \\
        
        \midrule
        \multicolumn{9}{l}{\textbf{Multi-Scale Feature-based Segmentation}}\\
        \midrule
        FCBFormer~\cite{Sanderson2022fcbformer} & 91.08\scriptsize$_{\textit{5.53}}$ & 88.48\scriptsize$_{\textit{8.44}}$  & 94.54\scriptsize$_{\textit{2.82}}$  & 93.63\scriptsize$_{\textit{3.49}}$ & 95.59\scriptsize$_{\textit{3.01}}$  & 89.83\scriptsize$_{\textit{5.71}}$ & 93.77\scriptsize$_{\textit{3.12}}$ & 94.08\scriptsize$_{\textit{2.05}}$  \\
        
        MSRF-Net~\cite{Srivastava2022msrfnet} & 93.71\scriptsize$_{\textit{2.90}}$ & 90.71\scriptsize$_{\textit{6.21}}$  & \textcolor{ib}{\textbf{96.69}}\scriptsize$_{\textit{0.67}}$  & 92.75\scriptsize$_{\textit{4.37}}$ & 94.74\scriptsize$_{\textit{3.86}}$  & 91.25\scriptsize$_{\textit{4.29}}$ & 94.61\scriptsize$_{\textit{2.28}}$ & 93.24\scriptsize$_{\textit{2.89}}$  \\
        
        CRCNet~\cite{Zhu2023crcnet} & 90.60\scriptsize$_{\textit{6.01}}$ & 89.38\scriptsize$_{\textit{7.54}}$  & 91.75\scriptsize$_{\textit{5.61}}$  & 89.94\scriptsize$_{\textit{7.18}}$ & 94.83\scriptsize$_{\textit{3.77}}$  & 94.85\scriptsize$_{\textit{0.69}}$ & 94.35\scriptsize$_{\textit{2.54}}$ & 92.63\scriptsize$_{\textit{3.50}}$  \\

        \midrule
        \multicolumn{9}{l}{\textbf{Attention-based Segmentation}}\\
        \midrule
        PraNet~\cite{Fan2020pranet} & 88.21\scriptsize$_{\textit{8.40}}$& 85.16\scriptsize$_{\textit{11.76}}$ & 87.24\scriptsize$_{\textit{10.12}}$ & 89.61\scriptsize$_{\textit{7.51}}$& 88.19\scriptsize$_{\textit{10.41}}$ & 85.88\scriptsize$_{\textit{9.66}}$& 87.73\scriptsize$_{\textit{9.16}}$& 88.32\scriptsize$_{\textit{7.81}}$ \\
        
        EGTransUNet~\cite{Pan2023egtransunet} & 93.19\scriptsize$_{\textit{3.42}}$ & 90.15\scriptsize$_{\textit{6.77}}$  & 93.02\scriptsize$_{\textit{4.34}}$  & 93.96\scriptsize$_{\textit{3.16}}$ & 95.12\scriptsize$_{\textit{3.48}}$  & 92.36\scriptsize$_{\textit{3.18}}$ & 94.34\scriptsize$_{\textit{2.55}}$ & \textcolor{red}{\textbf{96.27}}\scriptsize$_{\textit{0.14}}$  \\
        
        TGANet~\cite{Tomar2022tganet} & 88.58\scriptsize$_{\textit{8.03}}$ & 84.22\scriptsize$_{\textit{12.70}}$ & 89.74\scriptsize$_{\textit{7.62}}$  & 90.98\scriptsize$_{\textit{6.14}}$ & 94.98\scriptsize$_{\textit{3.62}}$  & 91.10\scriptsize$_{\textit{4.44}}$ & 94.81\scriptsize$_{\textit{2.08}}$ & 94.05\scriptsize$_{\textit{2.08}}$  \\
        
        CFHA-Net~\cite{Yang2023cfhanet} & 92.50\scriptsize$_{\textit{4.11}}$ & 90.30\scriptsize$_{\textit{6.62}}$  & 93.60\scriptsize$_{\textit{3.76}}$  & 92.86\scriptsize$_{\textit{4.26}}$ & 95.53\scriptsize$_{\textit{3.07}}$  & \textcolor{ib}{\textbf{95.54}}\scriptsize$_{\textit{0.35}}$ & 94.53\scriptsize$_{\textit{2.36}}$ & 95.01\scriptsize$_{\textit{1.12}}$  \\

        \midrule
        \multicolumn{9}{l}{\textbf{Multiscale Network with Spatial-enhanced Attention}}\\
        \midrule        
        MNet-SAt (our) & \textcolor{red}{\textbf{96.61}}\scriptsize$_{\textit{0.00}}$ & \textcolor{red}{\textbf{96.92}}\scriptsize$_{\textit{0.00}}$  & \textcolor{red}{\textbf{97.36}}\scriptsize$_{\textit{0.00}}$  & \textcolor{red}{\textbf{97.12}}\scriptsize$_{\textit{0.00}}$ & \textcolor{red}{\textbf{98.60}}\scriptsize$_{\textit{0.00}}$  & \textcolor{red}{\textbf{95.89}}\scriptsize$_{\textit{0.00}}$ & \textcolor{ib}{\textbf{96.89}}\scriptsize$_{\textit{0.00}}$ & \textcolor{ib}{\textbf{96.13}}\scriptsize$_{\textit{0.00}}$ \\
        \bottomrule
    \end{tabularx}    
\label{table:sotakscd}
\end{table*}

\begin{figure}[!ht]
  \centering
  \includegraphics[width=\linewidth]{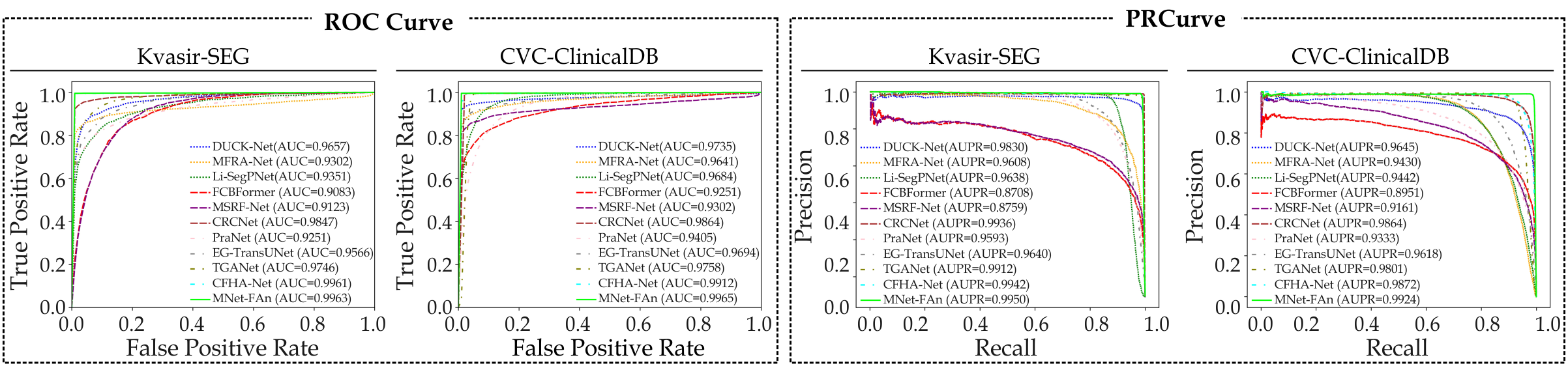}
  \caption{Analysis of ROC and PR curves on existing baselines}
  \label{fig:rocpr}
\end{figure}

\paragraph{Quantitative Evaluation}
\autoref{table:sotakscd} summarizes the quantitative results from a comprehensive evaluation of the Kvasir-SEG and CVC-ClinicalDB datasets. \cref{fig:rocpr} presents the corresponding Receiver Operating Characteristic (ROC) curves and Precision-Recall (PR) curves. Our proposed framework demonstrates superior performance, consistently achieving the highest DSC and IoU scores across both datasets. While Li-SegPNet~\cite{Sharma2023LiSegPNet} and EGTransUNet~\cite{Pan2023egtransunet} exhibit slightly higher Pre and Rec on CVC-ClinicalDB, these models lack the capability to capture multi-scale features as effectively as our framework. Moreover, the PR curves in \cref{fig:rocpr} further confirm that our framework demonstrates the best overall balance between Pre and Rec across both datasets. Compared to DUCK-Net~\cite{Dumitru2023ducknet}, our proposed framework demonstrates a 1.05\% and 1.45\% improvement in DSC on the Kvasir-SEG and CVC-ClinicDB datasets, respectively. Additionally, MNet-SAt exhibits comparable performance to MSRF-Net~\cite{Srivastava2022msrfnet} and MFRA-Net~\cite{Haider2023mfranet} in terms of Pre and Rec on the Kvasir-SEG dataset. On CVC-ClinicDB, our framework achieves DSC and IoU scores of 98.60\% and 95.89\%, respectively, outperforming DUCK-Net~\cite{Dumitru2023ducknet} by 1.45\% and CFHA-Net~\cite{Yang2023cfhanet} by 0.35\%. This quantitative analysis underscores the superior performance of MNet-SAt in polyp segmentation tasks.

\begin{figure*}
  \centering
  \includegraphics[width=\textwidth]{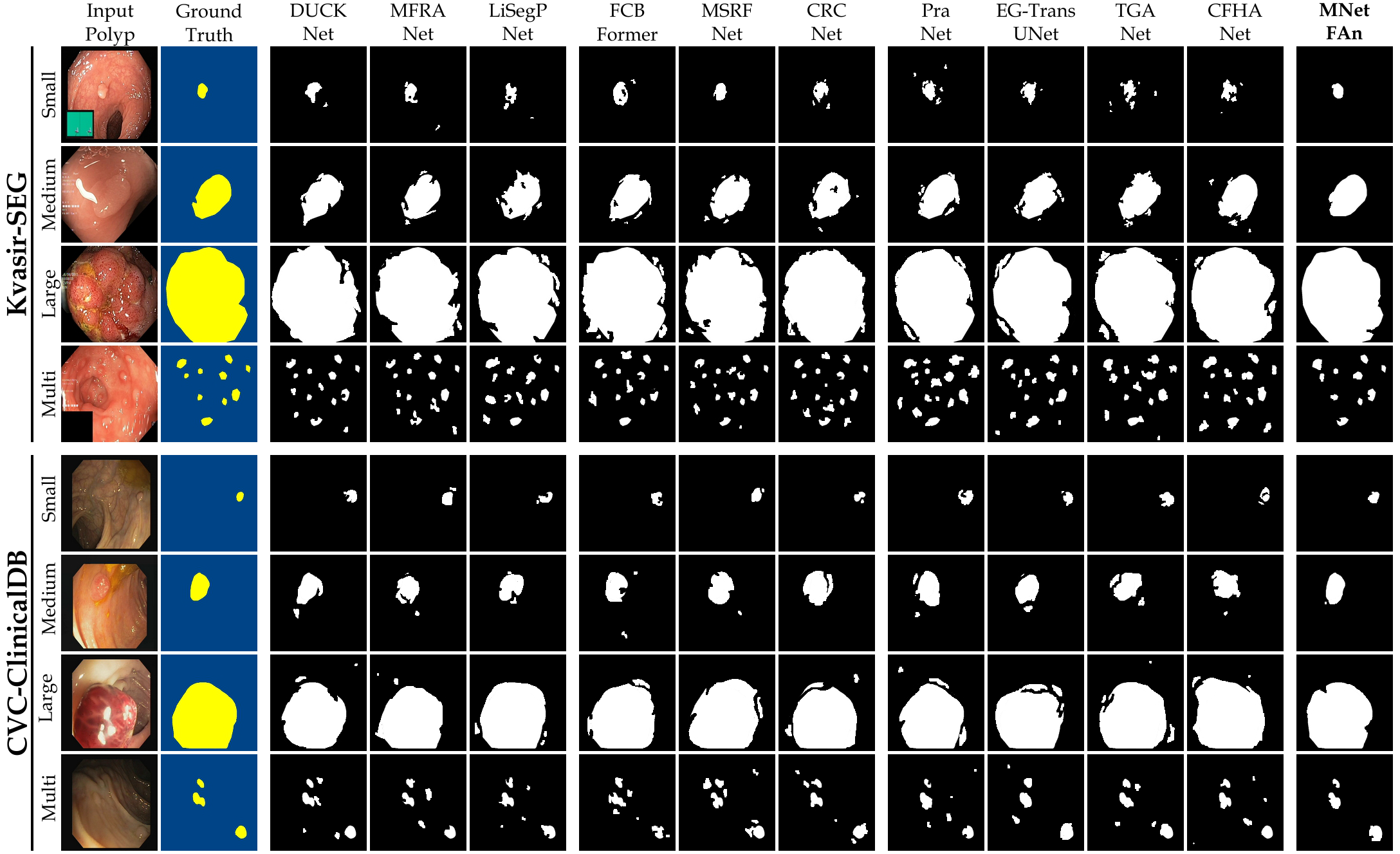}
  \caption{Qualitative comparison of MNet-SAt segmentation outputs with existing baselines}
  \label{fig:sotavis}
\end{figure*}

\paragraph{Qualitative Evaluation}
\cref{fig:sotavis} offers qualitative validation, showcasing MNet-SAt's superior polyp mask generation compared to baselines across four cases. This qualitative superiority aligns with the previously established quantitative results. In the case of small polyps (first and fifth rows), all baseline methods initially appear to produce satisfactory visual results. However, upon closer inspection, they fail to maintain fine boundary details. Our framework excels in preserving polyp anatomy and topology for medium and large polyps, significantly outperforming the baseline methods. Notably, even in multiple polyps, our framework successfully captures more polyps and approximates the ground truth more closely. The primary reason for this superior performance is the EGFE module, which effectively eliminates noise in conjunction with the HMAtt module, resulting in a few false positives. Our analysis reinforces MNet-SAt's effectiveness in handling challenging polyp scenarios (small, medium, large-scale, and multi-polyps) while suppressing non-regions of interest.

\subsubsection{Ablation Analysis}
\paragraph{Component-wise Ablation}
In this section, we systematically remove or modify individual components of MNet-SAt, keeping the rest of the framework unchanged. We denote ablated versions as \textit{MNet-SAt}$_{technique}$, where the subscript indicates the component under analysis, and ``w/o'' signifies without a specific component.

\parindent0pt\subparagraph{Effect of Encoder-Decoder Backbone:} To analyze the contributions of the encoder-decoder backbone, we conducted an ablation study with three variants: w/o EGFE, with EGFE only, and with both EGFE and Sobel operation (our proposed framework). The results presented in~\autoref{table:abl-ed} show that excluding EGFE leads to a degradation of DSC and IoU by 3.59\% and 4.05\%, respectively. Incorporating EGFE without the Sobel operation enhances edge pixel detection and improves DSC, IoU, Pre, and Rec by 1.29\%, 2.73\%, 1.52\%, and 1.33\%, respectively. These results confirm the efficacy of both the encoder-decoder backbone and the EGFE module integrated with Sobel operation in achieving superior polyp segmentation.

\begin{table}[!ht] \small
\caption{Ablation Study of Encoder-Decoder Backbone on MNet-SAt (Kvasir-SEG). \textbf{Bold}: Best results; $\nabla$: Absolute performance drop relative to MNet-SAt}
    \begin{tabularx}{\linewidth}{@{}l *4{C} @{}}
    
        \toprule        
        \textbf{Methods}
        &\textbf{DSC$_{|\nabla|}$} &\textbf{IoU$_{|\nabla|}$} &\textbf{Pre$_{|\nabla|}$} &\textbf{Rec$_{|\nabla|}$} \\
        
        \midrule        
        MNet-SAt$_{\text{w/o EGFE}}$ &
        93.02$_{\textit{3.59}}$ & 92.87$_{\textit{4.05}}$ & 93.55$_{\textit{3.81}}$ & 94.33$_{\textit{2.79}}$\\
        MNet-SAt$_{\text{EGFE + w/o Sobel}}$ &
        95.32$_{\textit{1.29}}$ & 94.19$_{\textit{2.73}}$ & 95.84$_{\textit{1.52}}$ & 95.79$_{\textit{1.33}}$\\
        
        \midrule        
        \textbf{MNet-SAt$_{\text{EGFE + Sobel}}$} &
        \textbf{96.61$_{\textit{0.00}}$} & \textbf{96.92$_{\textit{0.00}}$} & \textbf{97.36$_{\textit{0.00}}$} & \textbf{97.12$_{\textit{0.00}}$} \\ 
        
        \bottomrule
        \multicolumn{5}{l}{\scriptsize w/o stands for without, EGFE: Edge-Guided Feature Enrichment}
    \end{tabularx}    
\label{table:abl-ed}
\end{table}

\parindent0pt\subparagraph{Effect of MSFA Module:}
To gain insights into the MSFA module, we initially extract features at multiple scales $(\text{d}_\text{r} \in \text{1, 6, 12, 18})$, resulting in a DSC of 93.52\%, integrating the SE block after multi-scale feature extraction further improves DSC, IoU, Pre, and Rec. The SE block prioritizes informative features from extracted multi-scale extraction. Replacing the $3 \times 3$ convolution kernel with a $7 \times 7$ kernel in MSFA further improves performance, yielding a 3.09\% and 3.20\% increase in DSC and IoU, respectively, as highlighted in~\autoref{table:abl-msfa}. The results highlight the importance of the MSFA module in capturing crucial features for polyp segmentation.

\begin{table}[!ht] \small
\caption{Ablation Study of MSFA module on MNet-SAt (Kvasir-SEG). \textbf{Bold}: Best results; $\nabla$: Absolute performance drop relative to MNet-SAt}
    \begin{tabularx}{\linewidth}{@{}l *4{C} @{}}
    
        \toprule        
        \textbf{Methods}
        &\textbf{DSC$_{|\nabla|}$ } &\textbf{IoU$_{|\nabla|}$} &\textbf{Pre$_{|\nabla|}$} &\textbf{Rec$_{|\nabla|}$} \\
        
        \midrule        
        MNet-SAt$_{\text{$d_r$}}$ &
        93.52$_{\textit{3.09}}$ & 93.72$_{\textit{3.20}}$ & 93.58$_{\textit{3.78}}$ & 94.03$_{\textit{3.09}}$\\
        MNet-SAt$_{\text{$d_r$ + SE}}$ &
        95.16$_{\textit{1.45}}$ & 95.58$_{\textit{1.34}}$ & 96.14$_{\textit{1.22}}$ & 95.92$_{\textit{1.20}}$\\
        MNet-SAt$_{\text{$d_r$ + SE + Conv$_3$}}$ &
        96.07$_{\textit{0.54}}$ & 95.93$_{\textit{0.99}}$ & 96.74$_{\textit{0.62}}$ & 96.73$_{\textit{0.39}}$\\
        
        \midrule        
        \textbf{MNet-SAt$_{\text{$d_r$ + SE + Conv$_7$}}$} &
        \textbf{96.61$_{\textit{0.00}}$} & \textbf{96.92$_{\textit{0.00}}$} & \textbf{97.36$_{\textit{0.00}}$} & \textbf{97.12$_{\textit{0.00}}$} \\ 
        
        \bottomrule
        \multicolumn{5}{l}{\scriptsize \text{$d_r$}: dilation rate, SE: Squeeze \& Excitation, \text{$Conv_k$}: \text{$k \times k$} kernel Conv Operation}\\
    \end{tabularx}      
\label{table:abl-msfa}
\end{table}

\parindent0pt\subparagraph{Effect of SEAt Module:}
\autoref{table:abl-seat} summarizes the segmentation performance of the proposed MNet-SAt, which incorporates Layer Normalization (LN), a dilation rate of 2, and Global Average Pooling (GAP). In comparison to the variant without LN and a dilation rate of 1, the performance of the proposed MNet-SAt significantly improves, with an increase of 2.35\% and 3.41\% in DSC and IoU. Incorporating LN and increasing the dilation rate to two improves performance to 95.87\% and 95.79\% in DSC and IoU. Furthermore, including Global Average Pooling (GAP) within SEAt maintains spatial invariance and leads to further performance gains. These results underscore the role of the spatial enhanced attention module in improving polyp segmentation performance.

\begin{table}[!ht] \small
\caption{Ablation Study of SEAt module on MNet-SAt (Kvasir-SEG). \textbf{Bold}: Best results; $\nabla$: Absolute performance drop relative to MNet-SAt}
    \begin{tabularx}{\linewidth}{@{}l *4{C} @{}}
    
        \toprule        
        \textbf{Methods}
        &\textbf{DSC$_{|\nabla|}$ } &\textbf{IoU$_{|\nabla|}$} &\textbf{Pre$_{|\nabla|}$} &\textbf{Rec$_{|\nabla|}$} \\
        
        \midrule        
        MNet-SAt$_{\text{w/o LN + $d_r^1$}}$ &
        94.26$_{\textit{2.35}}$ & 93.51$_{\textit{3.41}}$ & 94.17$_{\textit{3.19}}$ & 94.12$_{\textit{3.00}}$\\
        MNet-SAt$_{\text{LN + $d_r^2$}}$ &
        95.87$_{\textit{0.74}}$ & 95.79$_{\textit{1.13}}$ & 95.38$_{\textit{1.98}}$ & 96.10$_{\textit{1.02}}$\\
        
        \midrule        
        \textbf{MNet-SAt$_{\text{ LN + $d_r^2$ + GAP}}$} &
        \textbf{96.61$_{\textit{0.00}}$} & \textbf{96.92$_{\textit{0.00}}$} & \textbf{97.36$_{\textit{0.00}}$} & \textbf{97.12$_{\textit{0.00}}$} \\ 
        
        \bottomrule
        \multicolumn{5}{l}{\scriptsize LN: Layer Normalization, \text{$d_r$}: dilation rate, GAP: Global Average Pooling} \\
    \end{tabularx}      
\label{table:abl-seat}
\end{table}

\parindent0pt\subparagraph{Effect of CE-ASPP:}
An ablation study in~\autoref{table:abl-ceaspp} evaluates the contributions of Multi-Scale Feature Fusion (MSF), Channel Compression (CC), and Inter-Channel Feature Fusion (ICF) within the CE-ASPP module. We observe that incorporating CC and MSF outperforms utilizing MSF alone but remains less effective than the MNet-SAt framework. Specifically, combining MSF with ICF (i.e., w/o CC) yields better results than combining CC with MSF. This enhanced performance is attributed to ICF, strengthening the framework's ability to capture complex dependencies between feature channels.

\begin{table}[!ht] \small
\caption{Ablation Study of CE-ASPP module on MNet-SAt (Kvasir-SEG). \textbf{Bold}: Best results; $\nabla$: Absolute performance drop relative to MNet-SAt}
    \begin{tabularx}{\linewidth}{@{}l *4{C} @{}}
    
        \toprule    
        \textbf{Methods}
        &\textbf{DSC$_{|\nabla|}$ } &\textbf{IoU$_{|\nabla|}$} &\textbf{Pre$_{|\nabla|}$} &\textbf{Rec$_{|\nabla|}$} \\
        
        \midrule 
        MNet-SAt$_{\text{MSF}}$ &
        94.21$_{\textit{2.40}}$ & 94.76$_{\textit{2.16}}$ & 94.39$_{\textit{2.97}}$ & 95.05$_{\textit{2.07}}$\\
        MNet-SAt$_{\text{CC + MSF}}$ &
        95.32$_{\textit{1.29}}$ & 95.41$_{\textit{1.51}}$ & 96.52$_{\textit{0.84}}$ & 95.87$_{\textit{1.25}}$\\
        MNet-SAt$_{\text{w/o CC + MSF + IF}}$ &
        95.89$_{\textit{0.72}}$ & 96.27$_{\textit{0.65}}$ & 97.08$_{\textit{0.28}}$ & 96.54$_{\textit{0.58}}$\\
        
        \midrule     
        \textbf{MNet-SAt$_{\text{CC + MSF + ICF}}$} &
        \textbf{96.61$_{\textit{0.00}}$} & \textbf{96.92$_{\textit{0.00}}$} & \textbf{97.36$_{\textit{0.00}}$} & \textbf{97.12$_{\textit{0.00}}$} \\ 
        
        \bottomrule
        \multicolumn{5}{l}{\scriptsize MSF: Multi-Scale Feature Fusion, CC: Channel Compression, ICF: Inter-Channel Feature Fusion} \\
    \end{tabularx}
\label{table:abl-ceaspp}
\end{table}

\paragraph{Analytical Ablation}
In this section, we conduct an ablation study to quantify the contribution of individual components within the MNet-SAt framework. The strategic addition of each module allows us to isolate its impact on the final segmentation performance. \autoref{table:abl-anat} shows that the Encoder-Decoder Backbone (EDB), utilizing EGFE units, delivers a baseline DSC of 93.50\%. Integrating the MSFA unit, which leverages variable dilation rates and SE blocks, significantly improves DSC by 1.24\%, highlighting its feature extraction effectiveness. Further integration of the SEAt module enhances spatial awareness within multi-scale aggregated features, leading to a 2.31\% DSC gain. Finally, the CE-ASPP module refines performance by resampling attentive features at different scales, resulting in an overall gain of 3.11\%. The last row of~\autoref{table:abl-anat} represents the complete MNet-SAt framework, demonstrating the cumulative benefit of all components.

\begin{table}[!ht] \small
\caption{Analytical Ablation of MNet-SAt (Kvasir-SEG). \textbf{Bold}: Best results; $\Delta$: Absolute performance gain relative to Encoder-Decoder backbone}
\centering
    \begin{tabularx}{\linewidth}{@{}P{1.2cm} P{1.2cm} P{1.2cm} P{2cm} *4{C} @{}}
        \toprule

        \textbf{EDB} &\textbf{MSFA} &\textbf{SEAt} &\textbf{CE-ASPP}  &\textbf{DSC$_{|\Delta|}$} &\textbf{IOU$_{|\Delta|}$} &\textbf{Pre$_{|\Delta|}$} &\textbf{Rec$_{|\Delta|}$} \\

        \midrule
        
        $\checkmark$ & $\times$ & $\times$ & $\times$ 
        & 93.50$_{\textit{0.00}}$ & 94.12$_{\textit{0.00}}$ & 94.07$_{\textit{0.00}}$ & 94.18$_{\textit{0.00}}$\\

        $\checkmark$ & $\checkmark$ & $\times$ & $\times$ 
        & 94.74$_{\textit{1.24}}$ & 95.19$_{\textit{1.07}}$ & 95.48$_{\textit{1.41}}$ & 95.22$_{\textit{1.04}}$ \\

        $\checkmark$ & $\checkmark$ & $\checkmark$ & $\times$ 
        & 95.81$_{\textit{2.31}}$ & 96.41$_{\textit{2.29}}$ & 96.71$_{\textit{2.64}}$ & 96.39$_{\textit{2.21}}$ \\
    
        $\pmb{\checkmark}$ & $\pmb{\checkmark}$ & $\pmb{\checkmark}$ & $\pmb{\checkmark}$
        & \textbf{96.61$_{\textit{3.11}}$} & \textbf{96.92$_{\textit{2.80}}$} & \textbf{97.36$_{\textit{3.29}}$} & \textbf{97.12$_{\textit{2.94}}$} \\
        \bottomrule

        \multicolumn{8}{l}{\scriptsize EDB: Encoder-Decoder Backbone, CE-ASPP: Channel-Enhanced Atrous Spatial Pyramid Pooling, MSFA: Multi-Scale}\\
        \multicolumn{8}{l}{\scriptsize Feature Aggregator, SEAt: Spatial-Enhanced Attention}\\
    \end{tabularx}
\label{table:abl-anat}
\end{table}

\subsubsection{Generalizability Evaluation}
To assess the generalizability of MNet-SAt, we conducted cross-training experimentation on the Kvasir-SEG and CVC-ClinicDB datasets. \autoref{table:cross-dataset} presents the results of cross-training evaluation on existing baselines. MNet-SAt demonstrates superior performance across most metrics. Specifically, when trained and tested on Kvasir-SEG and CVC-ClinicDB, it achieves the highest DSC of 91.52\%, IoU of 88.26\%, and Pre of 90.98\%. Similarly, when trained and tested on CVC-ClinicDB and Kvasir-SEG, MNet-SAt excels with a DSC of 92.28\%, Pre of 91.19\%, and Rec of 89.89\%. However, MFRA-Net~\cite{Haider2023mfranet} outperforms MNet-SAt in Rec when trained and tested on Kvasir-SEG and CVC-ClinicDB, and in IoU when trained and tested on CVC-ClinicDB and Kvasir-SEG.

\begin{table}[!b] \small
\caption{Generalization performance of MNet-SAt with existing baselines using 5-fold cross-validation. \textcolor{red}{\textbf{Red}} and \textcolor{ib}{\textbf{{blue}}} represent the best and second-best results}    
        \begin{tabularx}{\columnwidth}{@{} l *4{C} *4{C} @{}}
        \toprule

        \textbf{Train}        
        &\multicolumn{4}{c}{\textbf{Kvasir-SEG}} 
        &\multicolumn{4}{c}{\textbf{CVC-ClinicDB}} \\

        \textbf{Test}
        &\multicolumn{4}{c}{\textbf{CVC-ClinicDB}}        
        &\multicolumn{4}{c}{\textbf{Kvasir-SEG}} \\
        \cmidrule(l){1-1} \cmidrule(l){2-5} \cmidrule(l){6-9}         

        \textbf{Method}
        & \textbf{DSC} & \textbf{IoU} & \textbf{Pre} & \textbf{Rec}        
        & \textbf{DSC} & \textbf{IoU} & \textbf{Pre} & \textbf{Rec} \\
        
        \midrule    
        \multicolumn{9}{l}{\textbf{Encoder Decoder-based Segmentation}} \\
        \midrule        
        DUCK-Net~\cite{Dumitru2023ducknet}& \textcolor{ib}{\textbf{88.92}} & 86.11 & 87.65 & 86.23 & \textcolor{ib}{\textbf{91.81}} & 80.89 & 87.91 & 88.87 \\
        MFRA-Net~\cite{Haider2023mfranet}& 86.61 & 83.33 & 84.52 & \textcolor{red}{\textbf{90.72}} & 89.04 & \textcolor{red}{\textbf{87.90}} & 88.52 & 87.06 \\
        Li-SegPNet~\cite{Sharma2023LiSegPNet}& 83.78 & \textcolor{ib}{\textbf{86.62}} & 86.23 & 81.17 & 86.78 & 87.28 & 89.57 & 88.32 \\
        
        \midrule
        \multicolumn{9}{l}{\textbf{Multi-Scale Feature-based Segmentation}}\\
        \midrule
        FCBFormer~\cite{Sanderson2022fcbformer} & 85.00 & 81.41 & 87.80 & 86.32 & 89.39 & 84.61 & 87.47 & 87.76 \\
        MSRF-Net~\cite{Srivastava2022msrfnet} & 85.80 & 84.80 & 88.40 & 88.10 & 89.25 & 83.07 & 88.62 & 85.02 \\
        CRCNet~\cite{Zhu2023crcnet} & 85.95 & 83.09 & 84.88 & 83.02 & 87.50 & 87.62 & \textcolor{ib}{\textbf{90.02}} & 86.16 \\

        \midrule
        \multicolumn{9}{l}{\textbf{Attention-based Segmentation}}\\
        \midrule
        PraNet~\cite{Fan2020pranet} & 81.66 & 78.90 & 82.07 & 81.46 & 80.84 & 77.58 & 83.23 & 81.14 \\
        EGTransUNet~\cite{Pan2023egtransunet} & 85.99 & 83.39 & 84.99 & 87.65 & 87.54 & 83.78 & 89.18 & \textcolor{ib}{\textbf{89.04}} \\
        TGANet~\cite{Tomar2022tganet} & 83.51 & 78.05 & 84.28 & 84.37 & 87.40 & 84.76 & 89.13 & 86.74 \\
        CFHA-Net~\cite{Yang2023cfhanet} & 87.62 & 83.58 & \textcolor{ib}{\textbf{89.37}} & 86.60 & 88.45 & 86.82 & 88.52 & 88.36 \\

        \midrule
        \multicolumn{9}{l}{\textbf{Multiscale Network with Spatial-enhanced Attention}}\\
        \midrule        
        MNet-SAt (our) & \textcolor{red}{\textbf{91.52}} & \textcolor{red}{\textbf{88.26}} & \textcolor{red}{\textbf{90.98}} & \textcolor{ib}{\textbf{89.34}} & \textcolor{red}{\textbf{92.28}} & \textcolor{ib}{\textbf{87.62}} & \textcolor{red}{\textbf{91.19}} & \textcolor{red}{\textbf{89.89}} \\
        \bottomrule
    \end{tabularx}  
\label{table:cross-dataset}
\end{table}

Further analysis reveals that MNet-SAt outperforms DUCK-Net~\cite{Dumitru2023ducknet}, Li-SegPNet~\cite{Sharma2023LiSegPNet}, and CFHA-Net~\cite{Yang2023cfhanet} trained and tested on Kvasir-SEG and CVC-ClinicDB, achieving improvements of 2.5\%, 1.64\%, and 1.61\% in DSC, IoU, and Pre, respectively. Likewise, when trained and tested on CVC-ClinicDB and Kvasir-SEG, MNet-SAt surpasses DUCK-Net~\cite{Dumitru2023ducknet}, CRCNet~\cite{Zhu2023crcnet}, and EGTransUNet~\cite{Pan2023egtransunet} by margins of 0.47\%, 1.17\%, and 0.85\% in DSC, Pre, and Rec, respectively. Our framework achieves superior performance by integrating multi-scale feature learning with spatial attention. This unique combination surpasses existing methods limited to feature interactions at adjacent levels. Additionally, our encoder-decoder backbone enhances focus on polyp boundaries, improving contextual information capture.


\section{Conclusion}
In this research work, we propose a novel Multiscale Network with Spatial-enhanced Attention (MNet-SAt) for the segmentation of polyps in colonoscopy. MNet-SAt leverages a unique encoder-decoder architecture incorporating an edge-preserving unit to refine polyp boundaries. We enhance feature learning by channel-wise recalibrating and aggregating multi-scale features, capturing information across all scales. Subsequently, spatial-aware attention prioritizes global dependencies within these aggregated regions, highlighting salient features. The attentive features are then resampled at different scales for improved polyp representation. Extensive evaluations on Kvasir-SEG and CVC-ClinicDB datasets show that MNet-SAt outperforms existing methods. Future research will focus on adapting MNet-SAt for broader applicability in medical image segmentation, incorporating techniques such as self-supervised learning and knowledge distillation to enhance model robustness and generalizability for segmenting medical images.

\bibliographystyle{elsarticle-num} 

\bibliography{crefs}
\end{document}